# Performing Stance Detection on Twitter Data using Computational Linguistics Techniques

**Gourav G. Shenoy, Erika H. Dsouza, Sandra Kuebler**
Indiana University
Bloomington, IN, USA
`{goshenoy, ehdsouza, skuebler}@indiana.edu`

## Abstract

As humans, we can often detect from a person's utterances if he/she is in favor of or against a given target entity (topic, product, another person, etc). But from the perspective of a computer, we need means to automatically deduce the stance of the tweeter, given just the tweet text. In this paper, we present our results of performing stance detection on twitter data using a supervised approach. We begin by extracting bag-of-words to perform classification using TIMBL, then try and optimize the features to improve stance detection accuracy, followed by extending the dataset with two sets of lexicons - arguing, and MPQA subjectivity; next we explore the MALT parser and construct features using its dependency triples, finally we perform analysis using Scikit-learn Random Forest implementation.

## 1 Introduction

Stance detection for tweeter tweets involves detecting if the tweeter is in FAVOR or AGAINST a particular target which can be a person, a hot topic, etc. In this project, we have used various computational linguistic mechanisms to detect the stance.

Consider the target-tweet pair:
**Target:** Atheism
**Tweet:** *faith means believing in something without evidence like a fool.*

As humans, we can easily interpret the tweet to be in AGAINST of the target. We aim to build a model which would automatically detect the stance of the tweet. In order to detect the stance, the model's workflow should be able to identify the parts which are relevant and oftentimes may not be present in the entire tweet. For example, if a person is continuously giving quotes from the bible in his/her tweets, the model should realise that the person is against atheism and more towards believing in God. Due to this, the model is given a corpus for the target which would help in stance detection.

Stance detection is related to sentiment analysis but the two are different significantly. In sentiment analysis, given a tweet, the model has to decide if it is positive, negative or neutral. Whereas, in stance detection the model needs to detect if the tweet is in favour of the target and sometimes may not have the target explicitly mentioned in the tweet. Stance detection has a number of applications like text summarization, information retrieval and statistical analysis (For example, in the recent US presidential elections, many models were built for stance detection where the target was Donald Trump and many tweets predicted in favor of the newly elected President)

## 2 Data

The training data has 2913 total tweets and the test data has 1956 total tweets. The training data belongs to six targets: "Atheism", "Climate Change is a Real Concern", "Feminist Movement", "Hillary Clinton" and "Legalization of Abortion". The test data has an additional target "Donald Trump". No training data is explicitly provided for this target but a large set of tweets seem to be associated to it. Each of the tweets have a possible stance label:

- **FAVOR**: This label shows that the tweeter is in favor of the target.
- **AGAINST:** This label shows that the tweeter is not in favouritism of the target.
- **NONE**: This label is given when none of the above depictions can be made

## 3   Our Approach

We learnt a new hypothesis for each of the 6 targets (hillary, donald, feminism, legalization of abortion, climate, and atheism) by training a different model for each of them, with the default k-NN algorithm used by TIMBL. These methods, along with the results, are explained in detail below from section 3.1 to 3.8. What is interesting to note is that, for these models, we constructed the feature vectors using the following 3 strategies:

- Extracting bag-of-words of nouns, verbs, and adjectives for individual targets.
- Extracting bag-of-words with all words for individual targets.
- Extracting bag-of-words with nouns, verbs, and adjectives, along with the sentiment associated with the tweet, for individual targets - this is an optimization technique.

   We also tried tuning the parameters of TIMBL (k-NN algorithm) to analyse differences in accuracy. Subsequently, we extended our dataset by including the following subjectivity and arguing lexicons, and analysed the results. Finally we used the MALT parser to construct feature vectors by extracting dependency triples from the parsed train and test datasets, to perform our analysis.

### 3.1   Bag-of-Words as features

For feature creation using bag of words, the following steps were used:

1. train.csv and test.csv contained tweets belonging to different targets. We created separate train and test files for each target.

2. Used TnT to POS tag each of the tweets. TnT[1], the short form of *Trigrams n Tags*, is a very efficient statistical part-of-speech tagger that is trainable on different languages and virtually any tagset. The component for parameter generation trains on tagged corpora.

   The command used was:

   *$ tnt-para atheism_train.csv*
   *$ tnt tnt_data/penn_tnt a.csv > a.txt*

3. Created a python script that would group all the adjectives (**'JJ'**, **'JJR'**, **'JJS'**), nouns (**'NN'**, **'NNS'**, **'NNP'**) and verbs (**'VB'**, **'VBD'**, **'VBG'**, **'VBN'**, **'VBP'**, **'VBZ'**) that were tagged TnT parser.

4. Next, a python script created the feature vectors. So for every tweet, a 1 was inserted if the tweet had the corresponding noun, verbs or adjective and 0 otherwise. For example:

   *We Love God*

   *Tweet 1*   1   0   1
   *Tweet 2*   0   1   1

### 3.2   Classification using TIMBL default settings

**TiMBL** is an open source software package implementing several memory-based learning algorithms, among which IB1-IG, an implementation of k-nearest neighbor classification with feature weighting suitable for symbolic feature spaces. All implemented algorithms have in common that they store some representation of the training set explicitly in memory. During testing, new cases are classified by extrapolation from the most similar stored cases.

   The BOW were used as input to the TiMBL software using its default settings. The command used was:

*$ timbl -f  train.csv -t test.csv > output.txt*

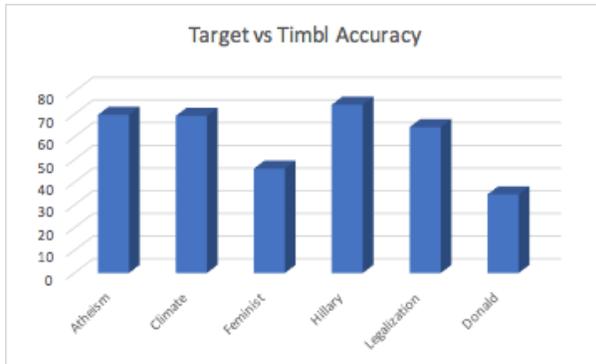

**Figure 1:** TiMBL accuracies for all targets with default settings

Here we observed that Timbl gave quite a good accuracy for the default setting (KNN with k=1) which was in the range of 45% (for Feminist Movement) to 74% (Hillary Clinton)

### 3.3 Tuning TIMBL parameters to improve accuracy

TIMBL accepts a parameter $k$ as command-line argument, which represents the number of nearest neighbors in k-NN algorithm it uses. TIMBL uses a default value of $k=1$ which will definitely overfit the training data, since with one nearest neighbor there are a large number of small clusters that are formed, and the model cannot generalize to new test data points.

Hence, we tried tuning this $k$ value to observe the alterations in accuracy. We noticed that by altering the $k$ value, the accuracy increases for higher values of $k$ until a point, after which it decreases and then after a point it remains constant.

Example: for category "atheism", the accuracy with default TIMBL settings is roughly 69%. But if we increase the k value beyond 11, the accuracy increases. We get highest accuracy of 76% for $k=13$, after which it decreases. Figure 2 shows the results for multiple $k$ values, for target *atheism*.

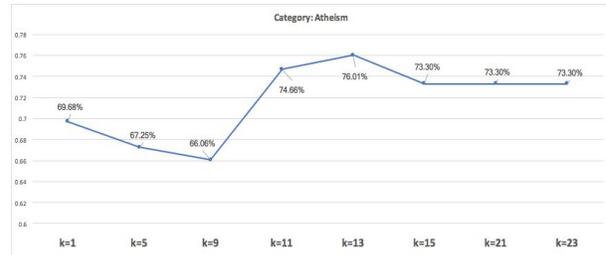

**Figure 2:** TIMBL accuracies for different $k$ values, target: *atheism*

We did this for all targets, learnt a completely new hypothesis with updated $k$ values for the algorithm and plotted the results; see Figure 3.

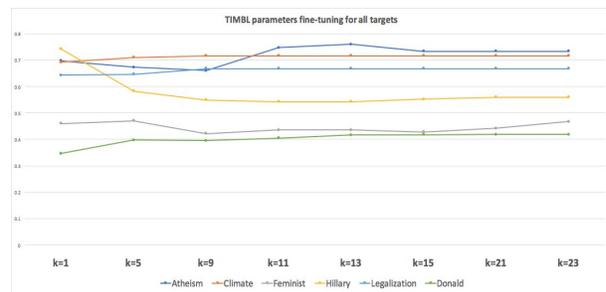

**Figure 3:** TIMBL accuracies for different $k$ values, all targets

### 3.4 Optimizing features

In the previous sections the feature vectors used to train the TIMBL model, purely consisted of only the bag-of-words extracted from the tweet text. We did not consider the *sentiment* associated with each tweet, which was also part of the dataset, while performing the learning (aka training). This additional *sentiment* information can be used to better understand the relationship between stance, sentiment, entity relationships, and textual inference.

We went ahead with our intuition that capturing these relationships via the *sentiment* feature would improve our stance detection accuracy, and we were absolutely correct. Although there wasn't a significant improvement in the accuracy, but we did observe a decent rise. Figure 4 compares the stance detection accuracies for all targets, with and without *sentiment* in the feature-vector.

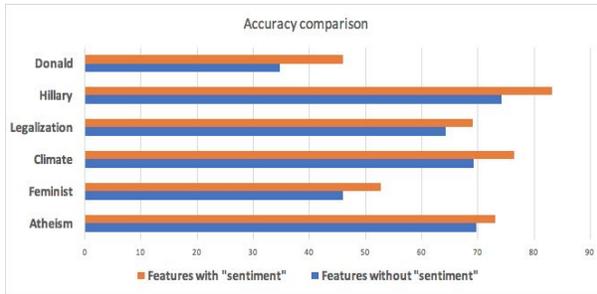

**Figure 4:** Accuracy comparison, features with and without *sentiment* data

Table 1 captures the accuracy numbers for the figure shown above. As you can see, there is a decent improvement in the accuracy for each target.

| Target | Features without "sentiment" | Features with "sentiment" |
|---|---|---|
| Atheism | 69.68% | 73.13% |
| Feminist | 45.96% | 52.66% |
| Climate | 69.23% | 76.41% |
| Legalization | 64.28% | 69.05% |
| Hillary | 74.16% | 83.17% |
| Donald | 34.65% | 45.98% |

**Table 1**: Accuracy comparison, features with and without *sentiment* information

### 3.5 Constructing Bag-of-Words using all words

The earlier analysis in section 3.2 showed Timbl results for features that were nouns, verbs and adjectives. In this section, we compare the Timbl results when we use all words.

We observe that when we use all words in our analysis, the accuracy drops by some percentage for each target except Donald Trump. The drop in accuracy is due to the fact that with large parameters, the 1-knn falls apart. See Figure 5.

### 3.6 Extending the dataset with MPQA subjectivity lexicons

Strategy for using Subjectivity Lexicons
For every word encountered in the twitter data-set (for each individual target), we look-up to see if that word is present in the subjectivity lexicon file.

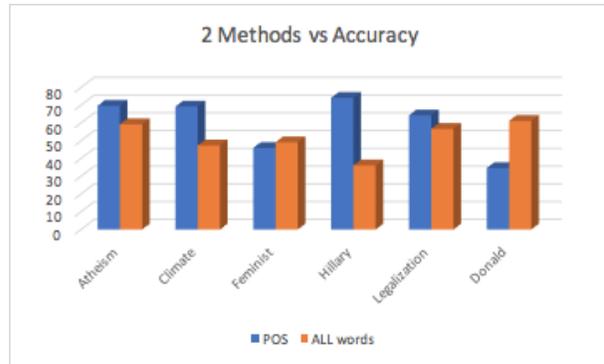

**Figure 5:** Accuracy comparison, features containing BOW with 3 POS tags v/s BOW with all words

Also since the lexicon file contains stemmed and unstemmed versions of a given word, we extract stems of the words in the twitter data-set and perform lookup. If there is a match, between a word in the tweets and the words in the lexicon file, we determine the "polarity" of that word given in the lexicon file. If the polarity is "positive" we associate a "+1" with the feature, and "-1" if the polarity is negative. If there is no match we associate a "0". This way, we provide weights to our feature vectors based on the polarity of the words and try to learn a model.

Reasoning for using strategy
By using weighted features, we are able to capture word-by-word meta-information about how strongly that feature contributes towards the stance provided in the training file. By extracting the stems of the tweeted words, we can cluster the words that are generated from a similar stem. Also, we used all words to construct our bag of words features. This way we hope to get more matches between the tweets and the words in the lexicon file.

Results
By using this strategy we noticed that our accuracy increases by a good amount (using default TIMBL settings). Table shows the accuracy comparison between BOW with/without MPQA lexicons for all targets.

| Target | 3 POS | MPQA |
|---|---|---|
| Atheism | 69.68% | 70.13% |
| Feminist | 45.96% | 59.65% |
| Climate | 69.23% | 71.59% |
| Legalization | 64.28% | 65.32% |
| Hillary | 74.16% | 54.57% |
| Donald | 34.65% | 32.11% |

Table 2: Accuracy comparison, with v/s without MPQA subjectivity lexicons

### 3.7 Extending dataset with arguing lexicons

The arguing lexicons consist of total twenty two *.tff* files; out of which seventeen files contain regular expressions representing "arguing", and five macro files. The strategy we employed to use these lexicon files is as follows:

- Check for occurrences of the regular expressions that match the extracted bag of words. Use macros wherever necessary.

- If there is a pattern match between the word and the regular expression representing an arguing category, we assign a weight of '1' to that feature, else assign '0'.

After running this strategy on the dataset provided, we observed that the regular expressions matched with only a few words in all categories. This also caused a drop in the stance detection accuracy as compared to the BOW strategy with MPQA lexicons. Table 3 asserts this accuracy comparison.

| Target | 3 POS | MPQA | Arguing |
|---|---|---|---|
| Atheism | 69.68% | 70.13% | 62.54% |
| Feminist | 45.96% | 59.65% | 56.88% |
| Climate | 69.23% | 71.59% | 59.26% |
| Legalization | 64.28% | 65.32% | 47.61% |
| Hillary | 74.16% | 54.57% | 49.95% |
| Donald | 34.65% | 32.11% | 30.72% |

**Table 3:** Accuracy comparison, BOW with 3 POS tags, MPQA and arguing lexicons

### 3.8 Constructing features using MALT parser

MaltParser is a system for data-driven dependency parsing, which can be used to induce a parsing model from treebank data and to parse new data using an induced model. The input that the MALTparser takes is in conll format. We used stanford-corenlp-full-2016-10-31 to parse the train and test tweets into conll format using command:

*$ java -cp "*" -Xmx2g edu.stanford.nlp.pipeline.StanfordCoreNLP -annotators tokenize,ssplit,pos,lemma,ner,parse,dcoref -file feminist_test.txt -outputFormat conll*

We then trained a parsing model for each target, using the command:

*$ java -jar maltparser-1.8.jar -c atheism -i ../stanford-corenlp-full-2016-10-31/legalization_train.conll -m learn -if myconll.xml*

**Note:** the myconll.xml is customised so that the maltparser can understand the conll syntax created by the stanford core nlp. The parsing model can now take in new sentences from the same language:

*$ java -jar maltparser-1.8.jar -c atheism -i ../stanford-corenlp-full-2016-10-31/legalization_test.conll -m parse -o out.conll*

Next, dependency triples were extracted from the data and used as features. For obtaining a "dependency trigram", we take the word, extract its head and the dependency label and create a "dependency trigram" feature out of that.

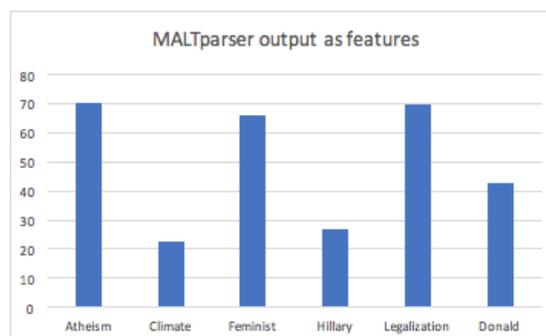

**Figure 6**: Accuracies with MALT parser output as features

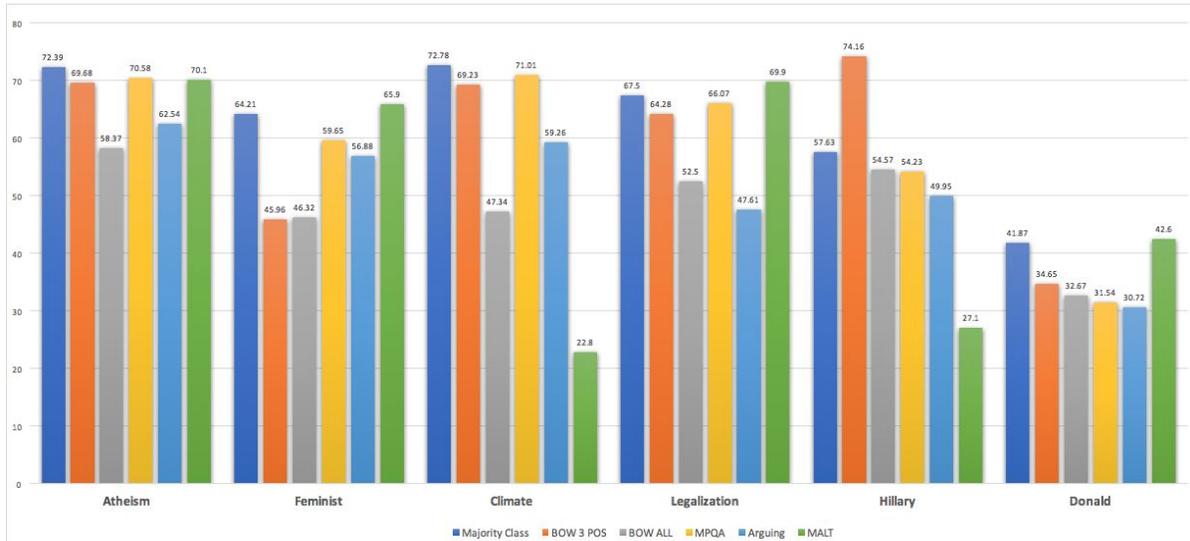

**Figure 7:** Accuracy comparison, all configurations - baseline (majority class), BOW with 3 POS tags, all words, MPQA and Arguing lexicons, and MALT parser; for all targets

We observe, that the performance using a dependency parser to create feature vectors doesn't improve the accuracy.

## 4 Scikit-learn Python Analysis

Scikit-learn provides a range of supervised and unsupervised learning algorithms via a consistent interface in Python. The library is built upon the SciPy (Scientific Python) that must be installed before you can use scikit-learn. In this section we will analyse the performance of Scikit-learn Random Forest implementation, for all the experiments we ran in section 3.1 to 3.8.

Random forests or random decision forests are an ensemble learning method for classification, regression and other tasks, that operate by constructing a multitude of decision trees at training time and outputting the class that is the mode of the classes (classification) or mean prediction (regression) of the individual trees. Random decision forests correct for decision trees' habit of overfitting to their training set.

### 4.1 Scikit-learn with Random Forest

We executed the Random Forest implementation of Scikit-learn with the same feature vectors used in sections 3.1 to 3.8 and recorded the accuracies for each target. We used this approach because the only difference between section 3 and 4, is the machine learning model that is used to learn the hypothesis. In section 3 we used TIMBL (and associated k-NN) to perform the learning, while in this section we will be using Random Forest (RF) which is provided by Scikit-learn python package. Sections 4.1.1 through 4.1.3 present results of the accuracies from Scikit-learn, for features made from bag-of-words, additional lexicons, and MALT parser output.

#### 4.1.1 Scikit RF with bag-of-words as features

We observed a similar behavior in stance detection accuracies of Scikit learn RF algorithm; features with 3 POS tags (noun, adjectives, verbs) provided better meaning to the vectors thereby resulting in higher overall accuracy in stance detection for each target. Table 4 asserts this observation.

*Note: We have used default parameters for Scikit-learn Random Forest. We haven't tuned the settings as of now.*

| Target | Majority Class | 3 POS | All Words |
|---|---|---|---|
| Atheism | 72.39% | 70.58% | 59.28% |
| Feminist | 64.21% | 48.07% | 49.12% |
| Climate | 72.78% | 67.45% | 47.34% |
| Legalization | 67.50% | 63.19% | 56.61% |
| Hillary | 57.63% | 56.27% | 36.07% |
| Donald | 41.87% | 35.36% | 61.17% |

Table 4: Scikit accuracy comparison, majority class v/s 3 POS tags, and all words in features

### 4.1.2 Scikit RF with additional lexicons

Again, after extending the dataset with MPQA and arguing lexicons, we used the same feature vectors for training as section 3.6, 3.7. The results showed similar behavior as expected; features with MPQA lexicons performed better than those with arguing. Table 5 lists the accuracies for reference.

| Target | MPQA | Arguing |
|---|---|---|
| Atheism | 73.13% | 65.11% |
| Feminist | 62.98% | 57.03% |
| Climate | 69.07% | 55.62% |
| Legalization | 62.66% | 47.87% |
| Hillary | 43.37% | 55.89% |
| Donald | 35.29% | 32.45% |

Table 5: Scikit accuracy comparison, MPQA v/s Arguing lexicons

### 4.1.3 Scikit RF with MALT output as features

We had a similar observation, the features generated from MALT parser output did little to help improve the accuracy for stance detection.

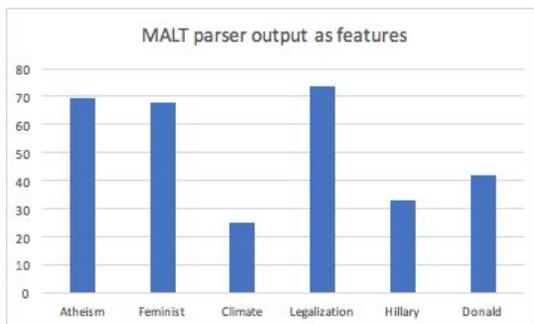

Figure 8: Scikit accuracies with MALT parser output as features

## 5 Conclusion

We discussed the various computational linguistics mechanisms we used to perform stance detection on the twitter data, for 6 different targets (hillary, donald, feminism, legalization of abortion, climate, and atheism). The basic idea we employed was to use machine learning to learn a new hypothesis for each of the 6 targets by training a different model for each of them, with the default k-NN algorithm by TIMBL. We began by extracting bag-of-words of nouns, adjectives, verbs (tagging performed using TnT) for individual targets and constructed feature-vectors to train our model. We performed classification using default TIMBL and observed that we achieved good accuracy with these features. We tried using all words as features, but noticed that the accuracy dropped drastically.

Tuning the TIMBL parameters, that is the $k$ value, helped improve accuracy; but we also observed a pattern where accuracy remains constant after a certain $k$ value. We optimized our features by including the *sentiment* of a tweet as a feature, and expectedly perceived an enhancement in the stance detection accuracy. By extending the dataset with two different set of lexicons, MPQA and Arguing; we noticed a decent increase in accuracy for MPQA, whereas due to low pattern matches with arguing lexs the accuracy dropped. Next we extracted dependency triples from the MALT parsed data (both train and test data) and used as features. We observed that the performance using a dependency parser to create feature vectors doesn't do much to improve the accuracy.

Finally, we repeated all our experiments using Scikit-learn Random Forest implementation. We observed similar behavior patterns using default settings for RF, but interestingly the accuracies were on the higher side this time.

### Acknowledgments

We would like to express great gratitude to Prof. Sandra Kuebler and associate instructor Ms. Atreyee Mukherjee for all their help, support and guidance throughout the project. This work would not have been possible without their help. We would also like to thank the School of Informatics

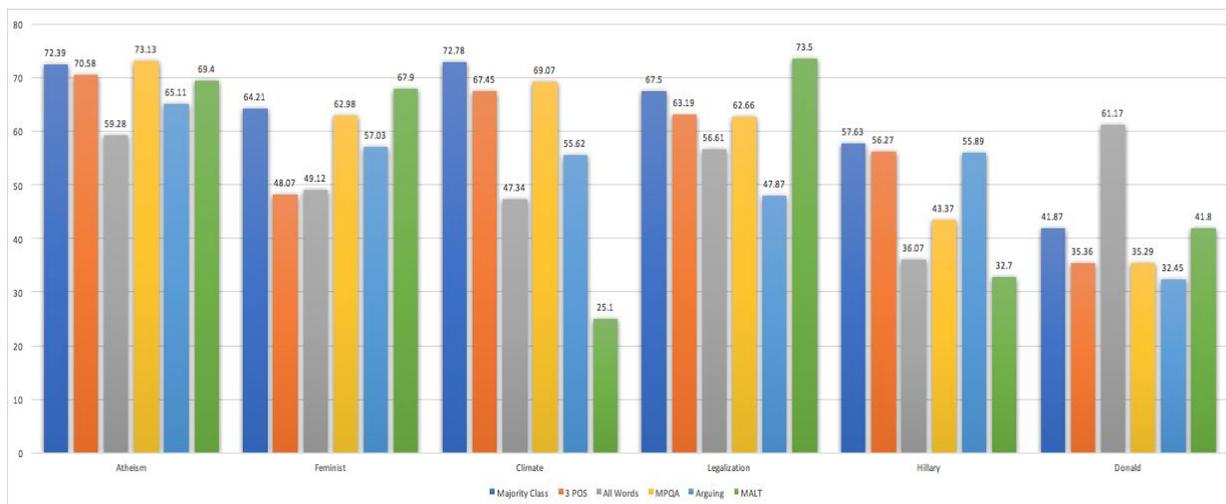

**Figure 9:** Scikit accuracy comparison, all configurations - baseline (majority class), BOW with 3 POS tags, all words, MPQA and Arguing lexicons, and MALT parser; for all targets

and Computing, and Department of Linguistics at Indiana University, for providing us with the infrastructure to work on this project. Finally, a big thank you to all class members for providing constant feedback and help with discussions as and when needed.